\documentclass[letterpaper, 10 pt, conference]{ieeeconf}      
\IEEEoverridecommandlockouts                              
\usepackage{algorithm}
\usepackage{algorithmic}
\usepackage{hyperref}
\usepackage{subfig}
\usepackage{amsmath}
\usepackage{amsfonts}
\usepackage{array}
\usepackage{todonotes}   
\usepackage{mathtools}
\title{
Vision-Based Multi-Task Manipulation for Inexpensive Robots Using End-To-End Learning from Demonstration
}

\author{Rouhollah Rahmatizadeh$^{1}$, Pooya Abolghasemi$^{1}$, Ladislau B{\"o}l{\"o}ni$^{1}$ and Sergey Levine$^{2}$
\thanks{$^{1}$ University of Central Florida, $^{2}$ University of California Berkeley}
}

\begin{document}

\maketitle
\thispagestyle{empty}
\pagestyle{empty}

\setlength{\belowdisplayskip}{1pt} \setlength{\belowdisplayshortskip}{1pt}
\setlength{\abovedisplayskip}{1pt} \setlength{\abovedisplayshortskip}{1pt}


\begin{abstract}

We propose a technique for multi-task learning from demonstration that trains the controller of a low-cost robotic arm to accomplish several complex picking and placing tasks, as well as non-prehensile manipulation. The controller is a recurrent neural network using raw images as input and generating robot arm trajectories, with the parameters shared across the tasks. The controller also combines VAE-GAN-based reconstruction with autoregressive multimodal action prediction. Our results demonstrate that it is possible to learn complex manipulation tasks, such as picking up a towel, wiping an object, and depositing the towel to its previous position, entirely from raw images with direct behavior cloning. We show that weight sharing and reconstruction-based regularization substantially improve generalization and robustness, and training on multiple tasks simultaneously increases the success rate on all tasks. 

\end{abstract}



\section{Introduction}
\label{sec:Introduction}

Autonomous open-world execution of a wide variety of manipulation skills is an important goal of robotic learning. Open-world manipulation must handle complex perception and a multitude of tasks. Of particular interest in this work is the performance of activities of daily living (ADLs): everyday tasks that able-bodied individuals perform with ease, but that can be a major challenge for a disabled or elderly person. Assistive robots performing ADLs need to operate in the uncontrolled environment of the user's home. To achieve maximum social impact, it would be desirable to develop perception and control methods that can use low-cost, imprecise hardware and readily-available sensory inputs such as camera images.

\begin{figure*}
  \begin{center}
    \includegraphics[width=1\textwidth]{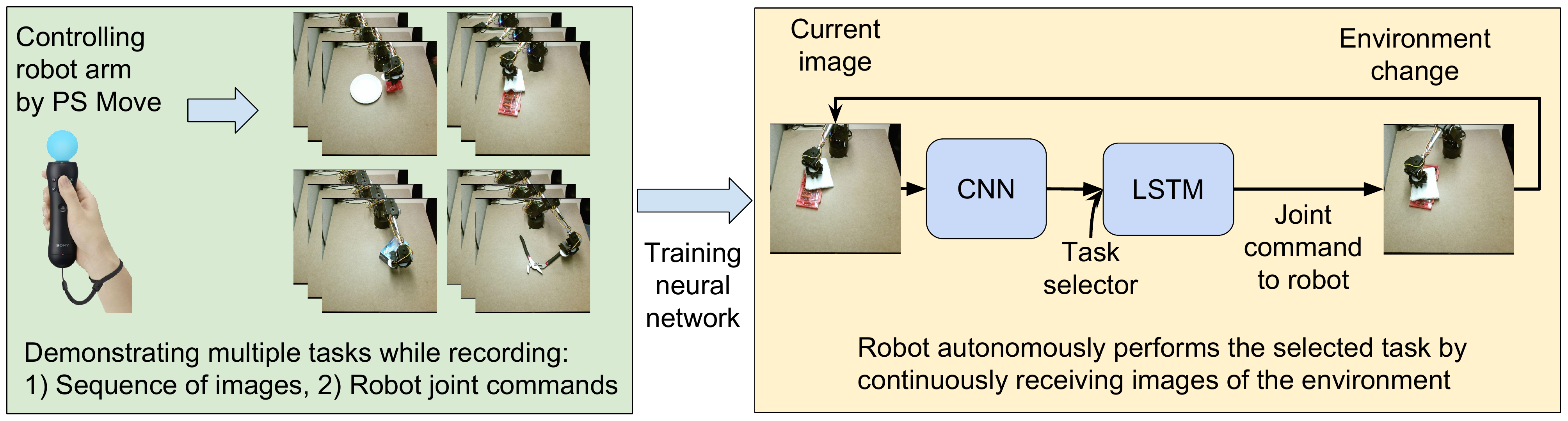}
  \end{center}
  \caption{An overview of our approach to multi-task learning }
  \label{fig:overview}
\end{figure*}

One approach to make robots autonomous is to hand-engineer a controller that is specific to a certain task. However, in the open-world settings we consider, hand-engineering a robust and resilient vision-based control strategy is exceptionally difficult. Learning from demonstration (LfD)
allows humans to demonstrate different tasks to the robot without having any knowledge about the robot's control or programming model. Because LfD can be performed using standard, efficient supervised learning methods, it requires orders of magnitude fewer interactions than reinforcement learning methods that learn from scratch. However, as the demonstrations take a lot of time from humans, reducing their number remains an important objective. In this paper, we propose an efficient learning approach that combines data from multiple tasks, as well as an auxiliary reconstruction loss, to learn a multi-task recurrent neural network policy from a tractable number of demonstrations. The policy takes as input images of the environment and a task selector one-hot vector, and predicts the joints of the robot in the next time-step. An overview of our approach is illustrated in Figure~\ref{fig:overview}.

Our work joins a number of recent papers that focus on learning robotic manipulation using deep neural networks. Within this field, our approach is differentiated by several features. We are training the robotic arm to execute multiple tasks, involving different objects ranging from rigid (a box), to jointed (pliers) and deformable (a towel). The tasks share all the network parameters and are distinguished only by a task selector one-hot vector. We are using a low-cost robot which does not have proprioception capabilities - all the information need to come from the vision component. 
Our controller uses an efficient autoregressive estimator, and regularization via image reconstruction to improve generalization and success rates. Finally, our approach is pure behavior cloning - we found that the tasks could be learned using the combination of techniques listed above, without the need of refining using dataset aggregation or reinforcement learning (although we are actively exploring whether such approaches can further improve the performance of the system). 

Our experimental results show that even with imprecise demonstrations and a low-cost manipulator, our method can perform a variety of manipulation tasks, including picking and placing and non-prehensile manipulation, can correct its mistakes, attempt tasks multiple times, and achieve high success rates. As far as we know, the learned behaviors\footnote{\href{https://youtu.be/AqQFzoVsJfA}{https://youtu.be/AqQFzoVsJfA}} represent the most complex autonomous task execution that had been successfully taught to robotic arms of this class using vision-based behavioral cloning. 
  
The primary contributions of this paper are as follows: 1)~We show that it is possible to control low-cost robotic platforms from monocular RGB images using a robust deep neural network based controller. 2)~We show that it possible to acquire complex manipulation skills, such as picking up a towel, wiping an object, and depositing the towel to its previous position entirely from raw images using direct behavior cloning
3)~We describe a novel model architecture that combines image reconstruction for regularization with autoregressive action prediction for sample-efficient learning of multimodal action distributions, and show that all these features contribute to the success of the learning
4)~We show that training a single network simultaneously on multiple tasks improves performance across all tasks. 

\section{Related Work}
\label{sec:RelatedWork}


A significant research area in contemporary robotics centers on moving from controlled environments and well-specified tasks to open world settings and tasks whose execution depends on the circumstances of the environment and preferences of the user. These settings favor approaches where the robot controller is learned rather than programmed. The two main approaches to learning robot behaviors are learning from demonstration (LfD) and reinforcement learning (RL). In LfD, the user demonstrates the desired behavior to the robot -- what the robot essentially learns is ``what the user would do in a situation like this?''. In case of RL, the robot tries to find the optimal policy by gathering experience in form of a reward function from its interactions with the environment. 

In recent years, LfD was successfully used in the field of robotics for applications as varied as autonomous helicopter maneuvers~\cite{abbeel2010autonomous}, playing table tennis~\cite{calinon2010learning}, object manipulation~\cite{pastor2009learning}, making coffee~\cite{sung_robobarista_2015}, grasping novel objects~\cite{kopicki2016one}, carrot grating~\cite{ureche2015task}, and eggplant cutting~\cite{lioutikov2016learning}. A major challenge in LfD is to extend these demonstrations to unseen situations~\cite{argall2009survey}. One obvious way to mitigate this problem is by acquiring a large number of demonstrations covering as many situations as possible. Some researchers proposed cloud-based and crowdsourced data collection techniques~\cite{kehoe2013cloud,forbes2014robot,crick2011human} or the use of simulation environments~\cite{fang2016learning}. Another direction is to use smaller number of demonstrations, but change the learning model to generalize better. One possible technique is to hand-engineer task-specific features~\cite{calinon2007learning,calinon2009handling}.

In contrast to LfD, which requires user demonstrations, RL approaches allow the robot to acquire different skills by itself ~\cite{kober2013reinforcement,peters2008reinforcement}. However, RL requires a reward signal that, in case of complex manipulation tasks, will be significantly delayed. The choice of LfD, RL or hybrid approaches is just one dimension in the design of learned robot controller. Both approaches might be applied to similar controllers, and in both cases, reducing the number of required demonstrations / experiments is a significant concern.


The recent successes of convolutional neural networks (CNNs) on computer vision tasks opened the possibility of creating end-to-end controllers, where a raw visual input (possibly augmented with robot proprioception), is translated by a single deep neural network into control signals. For instance,~\cite{levine2016end} used feed-forward neural networks to map a robot's visual input to control commands. The visual input is processed using a CNN to extract 2D feature points, then it is aggregated with the robot's current joint configuration, and fed into a feed-forward neural network. The resulting neural network will predict the next joint configuration of the robot. A similar neural network architecture is designed to address the grasping problem~\cite{pinto2015supersizing}. In~\cite{mayer2008system}, LSTMs are used for a robot to learn to autonomously tie knots after a pre-processing step to reduce the noise from the training data. Other examples of recent papers describing end-to-end-controllers include~\cite{agrawal2016learning,levine2016learning,lillicrap2015continuous}.

Another aspect of our work is multi-task learning. An early work considered learning a neural network policy on multiple related tasks with backpropagation~\cite{caruana1995learning}. It is possible to learn this kind of policies by considering both the state and the task as input to the policy~\cite{deisenroth2014multi}. Neural networks policies can be decomposed to different modules where there are some task-specific modules and some robot-specific modules~\cite{devin2016learning}. It is also shown that sharing the parameters of a neural network among different tasks not only improves the results, but also it is even better to train the model using the data of multiple related tasks instead of using the same amount of data from the original task~\cite{pinto2016learning}.

\section{Learning the Multi-Task Controller}
\label{sec:Method}

The overall flow of our method consists of (1) collecting a set of demonstrations for multiple tasks, (2) training a single deep recurrent neural network to emulate the user's behavior on all tasks simultaneously and (3) deploying the system, where the trained controller transforms raw camera perception to commands sent to the robot to perform the tasks.

\subsection{Task Demonstration and Data Collection} 

To collect demonstrations, we needed to enable human control of the robot using an intuitive teleoperation system. Low-cost robots do not have zero-gravity modes suitable for demonstration while remote control techniques based on a mouse or keyboard were found to be difficult to learn by assistive robotics users. The technique we used is for the robot end-effector to follow the user's hand, who can then use natural movements to demonstrate the task. We found two solutions for capturing the position and orientation of the hand: (a) using a Leap Motion controller and a ``naked'' hand\footnote{\href{https://www.youtube.com/watch?v=uAV5YcpuRu4}{https://youtu.be/uAV5YcpuRu4}}, and (b) using a Playstation Move controller\footnote{\href{https://www.youtube.com/watch?v=hSYta2T0Kg0}{https://youtu.be/hSYta2T0Kg0}}. Both the demonstration system and the arm are cheap (less than \$500 in total) and easily accessible by everyone. Therefore, this system can be deployed in people's home and they can teach arbitrary tasks to the robot based on their own preferences. 

During the demonstration we record the commands sent to the robot as well as $128\times128$ RGB images of the scene at a frequency of 33Hz. Then we down-sample the trajectories to reach a frequency of 4Hz. This way we create redundant trajectories with different starting point offsets~\cite{rahmatizadeh2016learning}. For instance, if we have a high frequency trajectory $\{ t_1, t_2, \ldots \}$, after down-sampling it by a factor of 8, we will have 8 trajectories $\{ t_1, t_9, t_{17}, \ldots \}$, $\{ t_2, t_{10}, t_{18}, \ldots \}$, etc. We found this data augmentation technique to be useful in regularizing the network.

\subsection{Neural Network Architecture}

\begin{figure*}
  \includegraphics[width=1\textwidth]{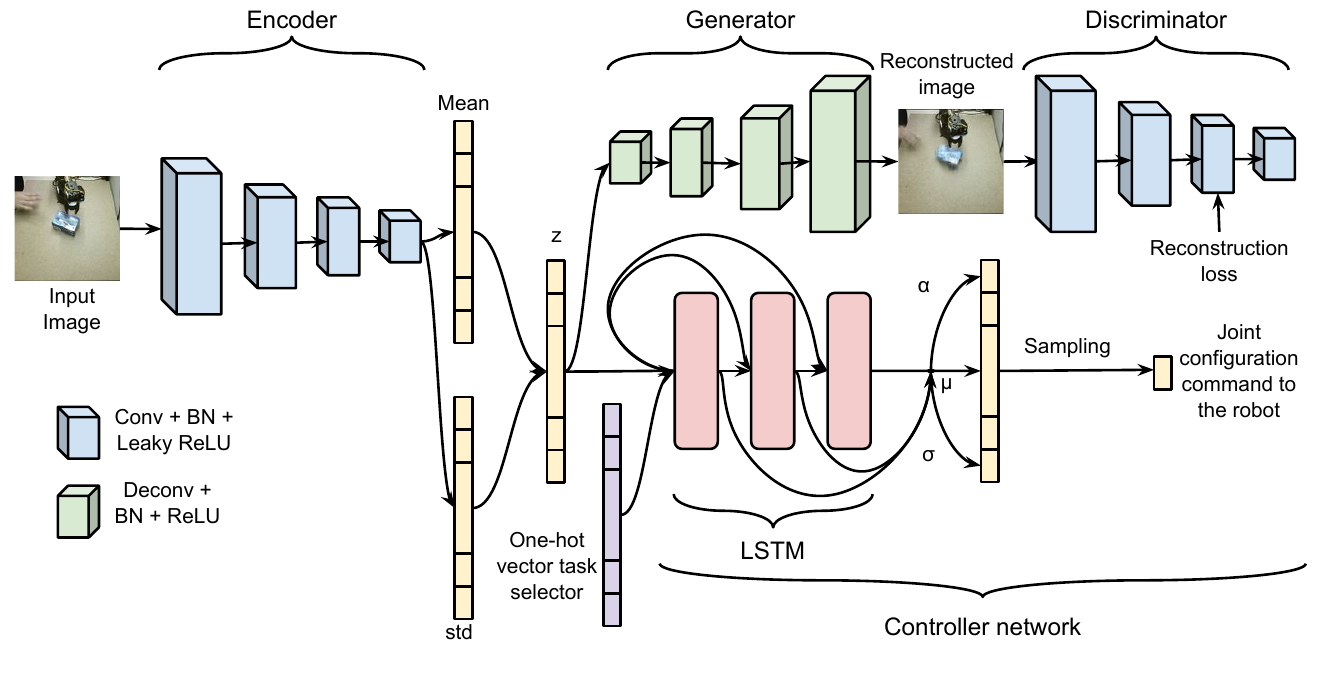}
  \caption{Our proposed architecture for multi-task robot manipulation learning. The neural network consists of a controller network that outputs joint commands based on a multi-modal autoregressive estimator and a VAE-GAN autoencoder
  that reconstructs the input image. The encoder is shared between the VAE-GAN autoencoder and the controller network and extracts some shared features that will be used for two tasks (reconstruction and controlling the robot). }
  \label{fig:network_architecture}
\end{figure*}

Our network architecture is illustrated in Figure~\ref{fig:network_architecture}. Convolutional layers augmented with batch normalization~\cite{ioffe2015batch} process the input images and map them to a low dimensional feature representation using the VAE-GAN approach~\cite{larsen2016autoencoding}. The VAE-GAN tries to encode input images based on the idea of Variational Autoencoders~\cite{kingma2013auto} and reconstruct realistic images based on the idea of Generative Adversarial Networks~\cite{goodfellow2014generative}. In this approach, a discriminator is added to the generator to discriminate the reconstructed images from the real images. However, instead of directly comparing the image pixels that causes uncertainty to appear in the form of blurriness, the comparison happens at the level of the extracted features of the real and reconstructed images after the third convolutional layer of the discriminator. On the bottom half of the Figure~\ref{fig:network_architecture}, we have a controller network where  the extracted visual features are combined with a task selector one-hot vector and fed into 3 layers of layer normalized~\cite{ba2016layer} LSTM~\cite{hochreiter1997long} to generate joint commands to control the robot. 

Most manipulation tasks can be solved in several distinct ways, and human demonstrators are often inconsistent about which solution they choose, even for the same task. A unimodal predictor, such as a Gaussian distribution, will average out dissimilar motions. By using a multi-modal predictor, we can capture different modes in the demonstrations without excessive averaging. However, simple multi-modal distributions such as mixtures of Gaussians~\cite{bishop1994mixture} provide a number of modes that scales linearly with the number of parameters. Instead, we are using a multi-modal autoregressive estimator similar to Neural Autoregressive Distribution Estimator (NADE)~\cite{larochelle2011neural}, to increase the number of modes the model can represent exponentially with the number of steps of the autoregressive model. While autoregressive estimators usually discretize the output, we use mixture of Gaussians to predict the entire probability distribution of the output, providing a rich and expressive class of distributions.

A closer look at the architecture shows that we have a VAE-GAN autoencoder that shares its encoder with the visual feature extractor of a controller network that sends commands to the robot. The autoencoder tries to fully reconstruct the images while the controller network will try to focus on some relevant features from the image such as the pose of the gripper and relevant objects. This competition/collaboration between these two networks will result in a more regularized visual feature extractor. This idea is similar to the semi-supervised learning with generative models~\cite{kingma2014semi} where they use a generative model via the VAE decoder and discriminative training via the action branch to improve sample efficiency. However, in contrast to this work, we observe an improvement in generalization simply from including the reconstruction objective, without including any additional unlabeled data.

Note that the extracted features from the encoder are in the form of a probability distribution that is encouraged to be close to the unit Gaussian by a KL-divergence penalty in the loss function.
The noise in the LSTM input caused by sampling from the encoded latent features helps to regularize the LSTM. In addition, we use dropout~\cite{srivastava2014dropout} with a probability of 0.5 to further avoid overfitting. 

\medskip

\subsection{Training the Network}

Our network is trained with backpropagation using two separate error signals. The first error signal trains the controller and is based on Mixture Density Networks (MDN)~\cite{bishop1994mixture}. The second error signal is based on the VAE-GAN architecture~\cite{larsen2016autoencoding} and is used to train the autoencoder network. 

\subsubsection{Mixture Density Network}
In the controller network, the output of the LSTM network is used to predict the parameters of a multi-modal mixture distribution. However, we do not predict all the outputs (joint configurations) at the same time-step of LSTM. Instead, we factor the j-dimensional distribution of joint configurations y(x) into a product of one-dimensional distributions, in this order: base, shoulder, elbow, ..., and gripper.
The probability distribution in this approach is modeled using a linear combination of Gaussian components of the form
\begin{equation}
p(y|x) = \sum_{i=1}^{m} \alpha_i(x) g_i(y|x)
\end{equation}
in which $\alpha_i(x)$ is the mixing coefficient, $g_i(y|x)$ is a multivariate Gaussian, and $m$ is a hyper-parameter that determines the number of components. At each time-step, the output is $y = y_t^j$, the current joint to be predicted and input $x$ is the encoded history of observations and predictions of the network before predicting the current joint. The Gaussian component is of the form
\begin{equation}
g_i(y|x) = \frac{1}{{ {(2\pi)}^{j/2}\sigma_i(x) }}\exp\left \{{{ - \frac{\|y - \mu_i(x)\|^2}{2\sigma_i(x)^2}}}\right \}
\end{equation}
where the vector $\mu_i(x)$ is the center of $i$th component. We do not calculate the full covariance matrices for each component, since this form of Gaussian mixture model is general enough to approximate any density function~\cite{mclachlan1988mixture}.

In the network architecture, we use skip connections from the input to all LSTM layers~\cite{graves2013generating}. Then the outputs of all LSTM layers are concatenated together and fully connected to another layer of width $3\times m$. This layer contains $m$ neurons each for $\mu_i(x)$, $\sigma_i(x)$ and $\alpha_i(x)$ respectively. To satisfy the constraint $\sum_{i=1}^{m} \alpha_i(x) = 1$, the corresponding neurons are passed through a softmax function. The neurons corresponding to the variances $\sigma_i(x)$ are passed through an exponential function and the neurons corresponding to the means $\mu_i(x)$ are used without any further changes. Finally, we can define the error of the controller network in terms of negative logarithm likelihood
\begin{equation}
E_\textit{controller} = -\ln \left \{ \sum_{i=1}^{m} \alpha_i(x) g_i(y|x) \right \}
\end{equation}


\subsubsection{VAE-GAN}

The autoencoder consists of three neural networks. The first network is an encoder that encodes the data sample $x$ to the latent representation $z \sim enc(x) = q(z|x)$.
Then the generator network decodes the latent representation $z$ and reconstructs the input image $\tilde{x} \sim gen(z) = p(x|z)$.
The last part of the autoencoder network is a GAN discriminator that takes a real image or a reconstructed image and tries to discriminate whether the given image is real or reconstructed.
\begin{equation}
P_\textit{real} = dis(x) \in [0,1]
\end{equation}
\begin{equation}
P_\textit{reconstruct} = dis(gen(z)) = 1 - P_\textit{real}
\end{equation}
The objective of the GAN is a min/max game to optimize the binary cross entropy described as:
\begin{equation}
E_\textit{GAN} = \log(dis(x)) + \log(1-dis(gen(z)))
\end{equation}
We impose a normal prior to the latent distribution $p(z)$ to regularize the encoder: 
\begin{equation}
E_\textit{prior} = D_\textit{KL}\left(q(z|x) \middle\| p(z)\right)
\end{equation}
To make the reconstructed image similar to the input image, we use mean squared error between the extracted features from those images in the third convolutional layer of the discriminator:
\begin{equation}
E_\textit{rec} = \textit{MSE}(dis_3(x), dis_3(\tilde{x}))
\end{equation}
Finally, the error of the autoencoder network can be described as the sum of errors formulated before:
\begin{equation}
E_\textit{AE} = E_\textit{prior} + E_\textit{rec} + E_\textit{GAN}
\end{equation}

\subsubsection{Implementation Details}
To train the controller and autoencoder networks, we alternate between them at each iteration with probability of 0.5 for the selection of each network. The parameters of the autoencoder are set and initialized based on the recommendations in~\cite{larsen2016autoencoding}, with the latent space size set to 256. All other parameters including the LSTM parameters are initialized uniformly between -0.08 to 0.08 following the recommendation in~\cite{sutskever2014sequence}. Each LSTM layer has 100 memory cells and is connected to a mixture of Gaussians with 50 components. 

The training of the network proceeds into two steps. We use this two-step training because the LSTM network needs a higher number of epochs to converge compared to the convolutional layers. In the first step we unroll and train the network using sequences of 5 time-steps and batch size of 100 examples for 5 epochs. In this step of training we use the Adam optimizer~\cite{kingma2014adam}. In the second step, we dump the extracted features from the convolutional layers to the disk and train only the LSTM layers using sequences of 50 time-steps and mini-batches of size 128 for 2000 epochs. In this step of training we switch to the RMSProp~\cite{tieleman2012lecture} optimizer with the initial learning rate of 0.005 and decay of 0.999. Also, the gradients are clipped in the range of [-1, 1] and the learning rate is decreased by a factor of 2 every 100 epochs.

\subsection{Executing the Policy at Test Time}
During the test time, the trained neural network controller generates robot joint commands in a loop by observing the environment. The LSTM predicts each joint one by one and when the predictions of all the joints are available, the robot takes an action, and another image is recorded and fed into the controller.
As we mentioned before, the user occasionally makes mistakes while demonstrating the task. However, it is desired that the robot does not repeat those mistakes. We can reduce the rate of mistakes by introducing some bias towards higher probability areas of the distribution while sampling from the probability distribution~\cite{graves2013generating}. While sampling, we use the new mixing coefficient
\begin{equation}
\alpha_i(x) = \frac{\exp(\alpha_i(x)(1+b))}{\exp(\sum_{i=1}^{m}\alpha_i(x)(1+b))}
\end{equation}
and standard deviation
\begin{equation}
\sigma_i(x) = \exp(\sigma_i(x)(1+b))
\end{equation}
where the bias parameter $b$ is a real number between 0 to 10. When $b=0$, there is no bias while $b=10$ introduces the maximum bias where only a point with maximum probability is chosen. We found $b=1$ to work well in our experiments.

\subsection{Why this Architecture?}

In this section we justify empirically the choices made in the design of the network. Later, in the experimental section we will also show that these choices significantly impact the performance. 

\medskip

\noindent {\em Why predict the entire probability distribution? } Many manipulation tasks can be solved in more than one way, and humans often choose randomly between the possible solutions. The simplest approach to behavior cloning would be to predict a deterministic joint command and use the mean squared error to measure the match between the predicted command and the user demonstrations. Unfortunately, if during demonstrations the user chooses different ways to solve the same problem, the  network will learn to predict the average of these commands. However, averaging between multiple solutions is usually not a correct solution. For instance, the robot arm might avoid an object from the left or from the right, but the average of these solutions is a collision with the object. Our approach models the entire multimodal probability distribution of solutions and samples its solution from this space. 

\medskip

\noindent {\em Why use recurrent neural networks?} As discussed above, humans might choose different solutions to a manipulation problem, but once decided, they need to stick to the chosen solution - they cannot choose between the ``avoid to the left'' or ``avoid to the right'' strategy at each time step. The memory present in a recurrent network helps the robot to remember and stick to a particular strategy similar to humans. Another benefit of the use of recurrent neural networks is situations where the current input image is not sufficient to predict the next action. For instance, sometimes the human demonstrator stops for a couple of time-steps to make sure the gripper is in the right place before grasping an object. In this case, a single image cannot determine whether to wait for another time-step or continue. The model needs to remember and count the number of time-steps that it has been waiting before closing the gripper. 

Another reason for using a recurrent neural network is that there will be timesteps where the model will fail to extract enough information from the current input to decide on what to do next. For instance, the manipulation object might be occluded or not encoded correctly due to the imperfection of the visual encoder. The LSTM recurrent neural networks are able to store this information and the controller can continue to act based on the network's memory until it regains the sight of the object. 


\medskip

\noindent {\em Why use an autoregressive density estimator?} By using this density estimator, we condition the prediction of each joint on the prediction of previous joints. Let us consider a situation in which an object is about to be grasped. The network needs to predict whether the gripper should be closed or not in the next time-step. If we do not condition the prediction of gripper on the prediction of other joints in the current time-step, the gripper might be closed before the end-effector is in a good grasping angle. Therefore, the grasp will fail. This idea of modeling the joint probability distribution of outputs by casting it as a product of conditional distributions is used in Pixel RNNs~\cite{oord2016pixel}, Neural Autoregressive Distribution Estimator (NADE)~\cite{larochelle2011neural}, and fully visible neural networks~\cite{neal1992connectionist,bengio1999modeling}.

\section{Experiments}
\label{sec:Experiments}


In order to evaluate our method, we consider several manipulation tasks that are frequent components of ADLs found in assistive robotics. The tasks involved the manipulation of objects on a table, by a robotic arm with a fixed base. The robot used was a 6-axis Lynxmotion AL5D robot with a two-finger gripper. The input video is recorded by a camera mounted facing the robot arm. This arrangement minimizes but does not completely eliminate the instances where the objects are occluded by the robot arm. The entire setup, including the arm, the camera, the Leap Motion and Playstation controllers used for demonstration, costs about \$500, making it affordable and accessible. The source code of the system is available as open source  at~\href{https://github.com/rrahmati/roboinstruct-2}{https://github.com/rrahmati/roboinstruct-2}. We encourage readers to reproduce or extend our experiements. 

\subsection{Manipulation Tasks}

We trained the robot to execute five manipulation tasks T1-5. All tasks involved manipulating objects, whose initial position and orientation was random, but within the reach of the robot arm. 


\noindent{\em T1: pick up a small bubble wrap and put it into a small plate}. This task is challenging since the robot needs to be very accurate while picking up the thin and deformable bubble wrap that often gets completely occluded by the arm during the grasp. If the bubble wrap is placed inside the plate, we count it as a success.

\noindent{\em T2: push a round plate to a specified area on the left side of the table}. This task is challenging because the robot needs to accurately detect the position of the plate and push at a point that moves the plate in the desired direction. In addition, if the plate ended up moving to an unexpected direction, the arm must push from a completely different contact point to fix the problem. If the plate is placed within an area with the radius of 2cm larger than the radius of the plate, we count it a success.

\noindent{\em T3: push a large box to a specific position and orientation close to the base of the robot arm}. This task is challenging because the robot needs to adjust the orientation of the box within a 20$^{\circ}$ error range and the position of the box within an area that is 2cm wider than the box in each direction. For instance, if the robot pushes the box further to the right such that it exits the target area, the arm has to circle around the box without colliding with it to push it from the other side.

\noindent{\em T4: close a set of open pliers and orient them parallel to the borders of the table}. In this task the convolutional layers need to detect the thin handles of the pliers so that the LSTM can decide where to push to accomplish the task. The pliers needs to be completely closed while 10$^{\circ}$ of error is acceptable for its final orientation. Note that in this task the initial orientation of the pliers is in a way that the handles are closer than its head to the base of the robot arm.

\noindent{\em T5: pick up a towel and rub a small screwdriver box to clean it}. We count the task as a success if the robot successfully picks up the towel, rubs the whole screwdriver box at least one time, and places the towel back on the table. (The quality of the cleaning was not considered in evaluating the success of this task). 

We collected 3 hours of demonstrations for each task, equivalent to 909, 495, 431, 428, 398 demonstrations for the tasks T1-5, respectively. We used 80\% of this data for training and kept the remaining 20\% for validation.


\subsection{Validating the autoencoder}

\begin{figure*}[ht]
  \includegraphics[width=1\textwidth]{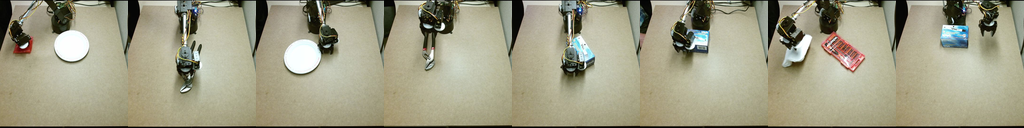}
   \includegraphics[width=1\textwidth]{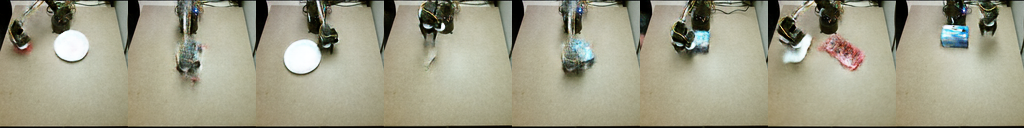}
  \caption{ Original input images to the network are shown on the top row. For each original image, the corresponding reconstructed image by the autoencoder is shown in the bottom row.}
  \label{fig:image_reconstruction}
\end{figure*}

Our network architecture uses an autoencoder to reduce the visual input to a latent space of 256 features. An advantage of this approach is that we can use the network to reconstruct the input images, and thus intuitively confirm that the features sufficiently capture the details of the scene to make possible the manipulation task. Figure~\ref{fig:image_reconstruction} shows the pairs of the original and reconstructed images for several images chosen from the manipulation tasks. We find that in most cases, the objects and the arm itself are captured and encoded reasonably well. Therefore, the LSTM has useful information to generate a trajectory to accomplish the task. 

\subsection{Results and Comparison of Network Architectures}
\label{sec:Results}

In the following set of experiments we aim to study both the performance achieved by the architecture and the degree to which the different network architectural choices contributed to this performance. In order to quantitatively evaluate the performance of our method, we allow the robot to try each task 25 times. If it cannot accomplish the task in a limited time (45 seconds for task 1, 60 seconds for tasks 2-4, and 75 seconds for task 5), we count the try as a failure, place the objects in a new random pose and repeat the experiment. Note that we do not stop the controller while we are resetting the experiment since this has also been the case during the demonstrations. It is interesting that the robot learned to go to the default state when it finishes the task since this was the preference of the user while demonstrating. Video clips of the demonstration process and the robot autonomously performing the five tasks can be watched online: \href{https://www.youtube.com/playlist?list=PL5i33tEH-MHfrXjj_Nekl0jyqgdJomg_a}{https://youtu.be/AqQFzoVsJfA}. 

Our network architecture contains a number of architectural choices: the use of a recurrent (LSTM) network, the use of the MDN as an error signal, the use of multi-task training, the use of an autoregressive estimator for joints, and the use of VAE-GAN for the reconstruction of the error signal. In our previous work on similar manipulation tasks, we had validated the superior performance of using LSTM instead of feedforward neural network and MDN instead of mean squared error loss for learning robot trajectories~\cite{rahmatizadeh2016learning}. Thus, in the experiments described in this paper, we will assume the existence of these components and evaluate the impact of adding the other components of the network in a comparison study. Table~\ref{table:task_results} shows the difference in performance of compared methods.

\begin{table}
{\footnotesize
\begin{tabular}{
p{4cm}p{.4cm}p{.4cm}p{.4cm}p{.4cm}p{.4cm} }  \hline \\[-.6em]

 \textbf{Method} & \textbf{T-1} & \textbf{T-2} &
 \textbf{T-3} &
 \textbf{T-4} &
 \textbf{T-5} \\ \\[-.8em]
 \hline \\[-.8em]
Single-task (no autoregressive) & 
 36\% & 16\% & 44\% & 16\% & 8\%\\  \\[-.8em]

Full network (no autoregressive) & 
16\% & 20\% & 52\% & 64\% & 20\% \\ \\[-.8em]

Full network (no VAE-GAN) &
12\% & 72\% & 56\% & 48\% & 16\%  \\ \\[-.8em]

Full network &
\textbf{76\%} & \textbf{80\%} & \textbf{88\%} & \textbf{76\%} & \textbf{88\%}  \\ \\[-.8em]

 \hline \\[-.8em]

\end{tabular}
}
\caption{Performance comparison of different methods. The numbers are the percentile rate of successfully accomplishing the tasks.}
\label{table:task_results}
\end{table}
 
\noindent{\em Single-task (no autoregressive).} For this set of experiments we used the network architecture from Figure~\ref{fig:network_architecture}, without the autoregressive estimator (all joints predicted at the same time). A separate network was trained on each task individually, without using the task selection vector. 

Although apparently a simpler task, training this network proved difficult since it overfits very easily. Our attempts to avoid or delay overfitting by increasing the dropout ratio or making the network smaller did not improve the results. The proposed model is very powerful and it does not have any assumption about the task or the shape of objects that are involved in each task. This is good since we can train the model on a wide variety of tasks. However, we need large number of demonstrations to successfully learn a single task. The controller trained using this approach generates some movements that are often similar to the demonstrations. However, they are not accurate enough to finish the task most of the time. 

\medskip

\noindent{\em Full network (no autoregressive).} In this series of experiments, we used the same network as above, but trained it on the data of all tasks and used the one-hot task selector vector to decide which task should be performed. For tasks T2-T5, this model worked significantly better than the single task model, while for task T1, it performed worse.


\medskip

\noindent{\em Full network (no VAE-GAN).} In this series of experiments, we augmented the previous approach with using the autoregressive estimator. However, we exclude the VAE/GAN that adds the reconstruction to the error signal. This way we can see if the reconstruction part of the network really helps in improving the learning. Based on our observation, it seems that the vision part of this model is not well trained especially in the tasks where the objects are smaller or occluded more often. The reason might be that the training examples are not enough to train the visual feature extractor of the controller. 

\medskip

\noindent{\em Full network.} Finally, in this set of experiments, we used all the architectural choices described in Section~\ref{sec:Method}. We find that for all five tasks, the full network outperforms the other choices, achieving performance of at least 76\% but as high as 88\% for tasks T-3 and T-5. 

\begin{figure*}
  \includegraphics[width=1\textwidth]{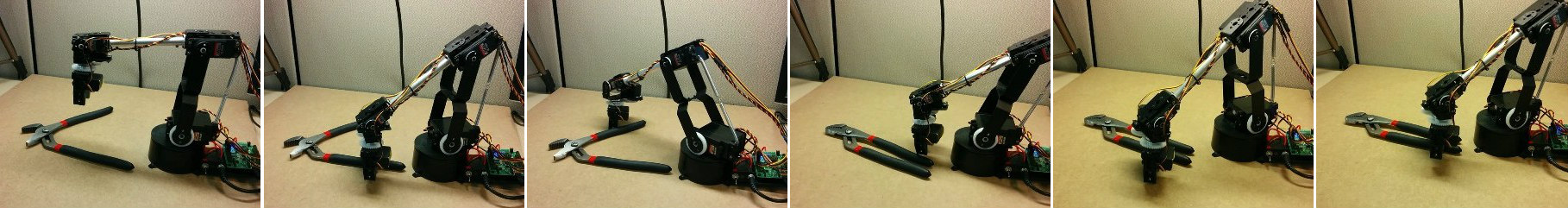}
  \caption{A sequence of images illustrating how the robot makes mistakes but fixes them on task T-4 (closing and orienting pliers). When the pliers are accidentally over-pushed, the robot moves to the other side of the pliers and pushes from the other direction to move the pliers back to the desired orientation.}
  \label{fig:mistake}
\end{figure*}

Beyond, the numerical task success rates, we also found that this controller generates very smooth trajectories that take a reasonable path in different situations. Interestingly, this controller has also learned to fix its own mistakes. An example of this mistake-fixing behavior is shown in Figure~\ref{fig:mistake}. 


\section{Conclusions}
\label{sec:Conclusions}

In this paper we proposed an approach to multi-task learning of manipulation tasks from user demonstrations. The approach takes as input images of the environment and outputs the next joint configuration of the robot. We showed that this approach works well on a low cost robot on basic manipulation tasks such as pushing and grasping. We also showed that the method can plan for longer tasks such as cleaning a small object using a towel. The proposed method is more sample-efficient compared to the single-task approach. This is because there is more data from the common patterns in different tasks that help the network to learn them easier.

While we demonstrate our approach on a set of five tasks, our results strongly suggest that the path toward robotic manipulators that can execute a wide variety of tasks lies in multi-task learning. Since our setup is inexpensive and accessible, we might imagine that, in the future, a large number of end-users could specify their own tasks to their own robots using our method. As more and more task demonstration data is collected, it can be used to train a single multi-task network. We expect that, as more tasks are added to the dataset, the number of individual demonstrations needed for each individual task will decrease, such that only a handful of additional demonstrations might be needed to allow the robot to perform each new behavior. Exploring this possibility is an exciting and promising direction for future work.

\noindent{\bf Acknowledgments:} This work had been supported by the
National Science Foundation under Grant Number
 IIS-1409823.

\bibliographystyle{ieeetr}
\bibliography{refs}

\begin{thebibliography}{10}

\bibitem{abbeel2010autonomous}
P.~Abbeel, A.~Coates, and A.~Y. Ng, ``Autonomous helicopter aerobatics through
  apprenticeship learning,'' {\em International Journal of Robotics Research
  (IJRR)}, 2010.

\bibitem{calinon2010learning}
S.~Calinon, F.~D'halluin, E.~L. Sauser, D.~G. Caldwell, and A.~G. Billard,
  ``Learning and reproduction of gestures by imitation,'' {\em IEEE Robotics \&
  Automation Magazine}, vol.~17, no.~2, pp.~44--54, 2010.

\bibitem{pastor2009learning}
P.~Pastor, H.~Hoffmann, T.~Asfour, and S.~Schaal, ``Learning and generalization
  of motor skills by learning from demonstration,'' in {\em IEEE International
  Conference on Robotics and Automation (ICRA)}, pp.~763--768, 2009.

\bibitem{sung_robobarista_2015}
J.~Sung, S.~H. Jin, and A.~Saxena, ``Robobarista: Object part-based transfer of
  manipulation trajectories from crowd-sourcing in 3d pointclouds,'' in {\em
  International Symposium on Robotics Research (ISRR)}, 2015.

\bibitem{kopicki2016one}
M.~Kopicki, R.~Detry, M.~Adjigble, R.~Stolkin, A.~Leonardis, and J.~L. Wyatt,
  ``One-shot learning and generation of dexterous grasps for novel objects,''
  {\em The International Journal of Robotics Research}, vol.~35, no.~8,
  pp.~959--976, 2016.

\bibitem{ureche2015task}
A.~L.~P. Ureche, K.~Umezawa, Y.~Nakamura, and A.~Billard, ``Task
  parameterization using continuous constraints extracted from human
  demonstrations,'' {\em IEEE Transactions on Robotics}, vol.~31, no.~6,
  pp.~1458--1471, 2015.

\bibitem{lioutikov2016learning}
R.~Lioutikov, O.~Kroemer, G.~Maeda, and J.~Peters, ``Learning manipulation by
  sequencing motor primitives with a two-armed robot,'' in {\em Intelligent
  Autonomous Systems}, pp.~1601--1611, Springer, 2016.

\bibitem{argall2009survey}
B.~D. Argall, S.~Chernova, M.~Veloso, and B.~Browning, ``A survey of robot
  learning from demonstration,'' {\em Robotics and autonomous systems},
  vol.~57, no.~5, pp.~469--483, 2009.

\bibitem{kehoe2013cloud}
B.~Kehoe, A.~Matsukawa, S.~Candido, J.~Kuffner, and K.~Goldberg, ``Cloud-based
  robot grasping with the {G}oogle object recognition engine,'' in {\em IEEE
  International Conference on Robotics and Automation (ICRA)}, pp.~4263--4270,
  2013.

\bibitem{forbes2014robot}
M.~Forbes, M.~J.-Y. Chung, M.~Cakmak, and R.~P. Rao, ``Robot programming by
  demonstration with crowdsourced action fixes,'' in {\em Second AAAI
  Conference on Human Computation and Crowdsourcing}, 2014.

\bibitem{crick2011human}
C.~Crick, S.~Osentoski, G.~Jay, and O.~C. Jenkins, ``Human and robot perception
  in large-scale learning from demonstration,'' in {\em International
  conference on Human-robot interaction}, pp.~339--346, ACM, 2011.

\bibitem{fang2016learning}
Z.~Fang, G.~Bartels, and M.~Beetz, ``Learning models for constraint-based
  motion parameterization from interactive physics-based simulation,'' in {\em
  IEEE/RSJ International Conference on Intelligent Robots and Systems (IROS)},
  pp.~4005--4012, 2016.

\bibitem{calinon2007learning}
S.~Calinon, F.~Guenter, and A.~Billard, ``On learning, representing, and
  generalizing a task in a humanoid robot,'' {\em IEEE Transactions on Systems,
  Man, and Cybernetics}, vol.~37, no.~2, pp.~286--298, 2007.

\bibitem{calinon2009handling}
S.~Calinon, F.~D'halluin, D.~G. Caldwell, and A.~Billard, ``Handling of
  multiple constraints and motion alternatives in a robot programming by
  demonstration framework.,'' in {\em IEEE International Conference on Humanoid
  Robots (Humanoids)}, pp.~582--588, Citeseer, 2009.

\bibitem{kober2013reinforcement}
J.~Kober, J.~A. Bagnell, and J.~Peters, ``Reinforcement learning in robotics: A
  survey,'' {\em The International Journal of Robotics Research}, vol.~32,
  no.~11, pp.~1238--1274, 2013.

\bibitem{peters2008reinforcement}
J.~Peters and S.~Schaal, ``Reinforcement learning of motor skills with policy
  gradients,'' {\em Neural networks}, vol.~21, no.~4, pp.~682--697, 2008.

\bibitem{levine2016end}
S.~Levine, C.~Finn, T.~Darrell, and P.~Abbeel, ``End-to-end training of deep
  visuomotor policies,'' {\em Journal of Machine Learning Research}, vol.~17,
  no.~39, pp.~1--40, 2016.

\bibitem{pinto2015supersizing}
L.~Pinto and A.~Gupta, ``Supersizing self-supervision: Learning to grasp from
  50k tries and 700 robot hours,'' in {\em IEEE International Conference on
  Robotics and Automation (ICRA)}, pp.~763--768, 2016.

\bibitem{mayer2008system}
H.~Mayer, F.~Gomez, D.~Wierstra, I.~Nagy, A.~Knoll, and J.~Schmidhuber, ``A
  system for robotic heart surgery that learns to tie knots using recurrent
  neural networks,'' in {\em IEEE/RSJ International Conference on Intelligent
  Robots and Systems (IROS)}, pp.~543--548, 2006.

\bibitem{agrawal2016learning}
P.~Agrawal, A.~Nair, P.~Abbeel, J.~Malik, and S.~Levine, ``Learning to poke by
  poking: Experiential learning of intuitive physics,'' {\em arXiv preprint
  arXiv:1606.07419}, 2016.

\bibitem{levine2016learning}
S.~Levine, P.~Pastor, A.~Krizhevsky, and D.~Quillen, ``Learning hand-eye
  coordination for robotic grasping with deep learning and large-scale data
  collection,'' {\em arXiv preprint arXiv:1603.02199}, 2016.

\bibitem{lillicrap2015continuous}
T.~P. Lillicrap, J.~J. Hunt, A.~Pritzel, N.~Heess, T.~Erez, Y.~Tassa,
  D.~Silver, and D.~Wierstra, ``Continuous control with deep reinforcement
  learning,'' {\em arXiv preprint arXiv:1509.02971}, 2015.

\bibitem{caruana1995learning}
R.~Caruana, ``Learning many related tasks at the same time with
  backpropagation.,'' {\em Advances in neural information processing systems},
  pp.~657--664, 1995.

\bibitem{deisenroth2014multi}
M.~P. Deisenroth, P.~Englert, J.~Peters, and D.~Fox, ``Multi-task policy search
  for robotics,'' in {\em IEEE International Conference on Robotics and
  Automation (ICRA)}, pp.~3876--3881, IEEE, 2014.

\bibitem{devin2016learning}
C.~Devin, A.~Gupta, T.~Darrell, P.~Abbeel, and S.~Levine, ``Learning modular
  neural network policies for multi-task and multi-robot transfer,'' {\em arXiv
  preprint arXiv:1609.07088}, 2016.

\bibitem{pinto2016learning}
L.~Pinto and A.~Gupta, ``Learning to push by grasping: Using multiple tasks for
  effective learning,'' {\em arXiv preprint arXiv:1609.09025}, 2016.

\bibitem{rahmatizadeh2016learning}
R.~Rahmatizadeh, P.~Abolghasemi, A.~Behal, and L.~B{\"o}l{\"o}ni, ``Learning
  real manipulation tasks from virtual demonstrations using {LSTM},'' {\em
  arXiv preprint arXiv:1603.03833}, 2016.

\bibitem{ioffe2015batch}
S.~Ioffe and C.~Szegedy, ``Batch normalization: Accelerating deep network
  training by reducing internal covariate shift,'' {\em arXiv preprint
  arXiv:1502.03167}, 2015.

\bibitem{larsen2016autoencoding}
A.~B.~L. Larsen, S.~K. S{\o}nderby, H.~Larochelle, and O.~Winther,
  ``Autoencoding beyond pixels using a learned similarity metric,'' in {\em
  International Conference on Machine Learning}, pp.~1558--1566, 2016.

\bibitem{kingma2013auto}
D.~P. Kingma and M.~Welling, ``Auto-encoding variational bayes,'' {\em arXiv
  preprint arXiv:1312.6114}, 2013.

\bibitem{goodfellow2014generative}
I.~Goodfellow, J.~Pouget-Abadie, M.~Mirza, B.~Xu, D.~Warde-Farley, S.~Ozair,
  A.~Courville, and Y.~Bengio, ``Generative adversarial nets,'' in {\em
  Advances in neural information processing systems}, pp.~2672--2680, 2014.

\bibitem{ba2016layer}
J.~L. Ba, J.~R. Kiros, and G.~E. Hinton, ``Layer normalization,'' {\em arXiv
  preprint arXiv:1607.06450}, 2016.

\bibitem{hochreiter1997long}
S.~Hochreiter and J.~Schmidhuber, ``Long short-term memory,'' {\em Neural
  computation}, vol.~9, no.~8, pp.~1735--1780, 1997.

\bibitem{bishop1994mixture}
C.~M. Bishop, ``Mixture density networks,'' {\em Technical Report}, 1994.

\bibitem{larochelle2011neural}
H.~Larochelle and I.~Murray, ``The neural autoregressive distribution
  estimator.,'' in {\em AISTATS}, vol.~1, p.~2, 2011.

\bibitem{kingma2014semi}
D.~P. Kingma, S.~Mohamed, D.~J. Rezende, and M.~Welling, ``Semi-supervised
  learning with deep generative models,'' in {\em Advances in Neural
  Information Processing Systems}, pp.~3581--3589, 2014.

\bibitem{srivastava2014dropout}
N.~Srivastava, G.~Hinton, A.~Krizhevsky, I.~Sutskever, and R.~Salakhutdinov,
  ``Dropout: A simple way to prevent neural networks from overfitting,'' {\em
  The Journal of Machine Learning Research}, vol.~15, no.~1, pp.~1929--1958,
  2014.

\bibitem{mclachlan1988mixture}
G.~J. McLachlan and K.~E. Basford, ``Mixture models. inference and applications
  to clustering,'' {\em Statistics: Textbooks and Monographs, New York:
  Dekker}, vol.~1, 1988.

\bibitem{graves2013generating}
A.~Graves, ``Generating sequences with recurrent neural networks,'' {\em arXiv
  preprint arXiv:1308.0850}, 2013.

\bibitem{sutskever2014sequence}
I.~Sutskever, O.~Vinyals, and Q.~V. Le, ``Sequence to sequence learning with
  neural networks,'' in {\em Advances in neural information processing systems
  (NIPS)}, pp.~3104--3112, 2014.

\bibitem{kingma2014adam}
D.~Kingma and J.~Ba, ``Adam: A method for stochastic optimization,'' {\em arXiv
  preprint arXiv:1412.6980}, 2014.

\bibitem{tieleman2012lecture}
T.~Tieleman and G.~Hinton, ``Lecture 6.5-rmsprop: Divide the gradient by a
  running average of its recent magnitude,'' {\em COURSERA: Neural Networks for
  Machine Learning}, vol.~4, 2012.

\bibitem{oord2016pixel}
A.~V.~d. Oord, N.~Kalchbrenner, and K.~Kavukcuoglu, ``Pixel recurrent neural
  networks,'' in {\em International Conference on Machine Learning},
  pp.~1747--1756, 2016.

\bibitem{neal1992connectionist}
R.~M. Neal, ``Connectionist learning of belief networks,'' {\em Artificial
  intelligence}, vol.~56, no.~1, pp.~71--113, 1992.

\bibitem{bengio1999modeling}
Y.~Bengio and S.~Bengio, ``Modeling high-dimensional discrete data with
  multi-layer neural networks.,'' in {\em Advances in neural information
  processing systems (NIPS)}, vol.~99, pp.~400--406, 1999.

\end{thebibliography}

\end{document}